\documentclass[letterpaper, 10 pt, conference]{ieeeconf}  

\IEEEoverridecommandlockouts                              

\usepackage{bm, amsmath, amssymb, gensymb, optidef}
\usepackage{algpseudocode, algorithm}
\usepackage{graphicx, subcaption}
\usepackage[switch]{lineno}
\graphicspath{ {figures/} }
\usepackage{tikz}
\usepackage{cite}

\makeatletter
\let\NAT@parse\undefined
\makeatother
\usepackage[hyphens]{url}
\usepackage{breakurl}
\usepackage{hyperref}
\definecolor{dark_green}{rgb}{0.0, 0.7, 0.0}
\hypersetup{
    colorlinks=true,
    citecolor=dark_green,
    urlcolor=blue,
    linkcolor=blue, 
}

\DeclareMathOperator*{\argmin}{argmin}

\title{\LARGE \bf
Relative Contact Velocity-Controlled Hand-Object Mechanism\\for Dexterous Tool Manipulation
}

\author{
Sunyu Wang, Jean Oh, and Nancy S. Pollard
\thanks{The authors are with the Robotics Institute at Carnegie Mellon University, Pittsburgh, USA. Corresponding author's contact: \textcolor{red}{sunyuw@andrew.cmu.edu}}
}

\begin{document}
\makeatletter
\let\@oldmaketitle\@maketitle
\renewcommand{\@maketitle}{\@oldmaketitle
\centering
  \includegraphics[width=0.99\linewidth]{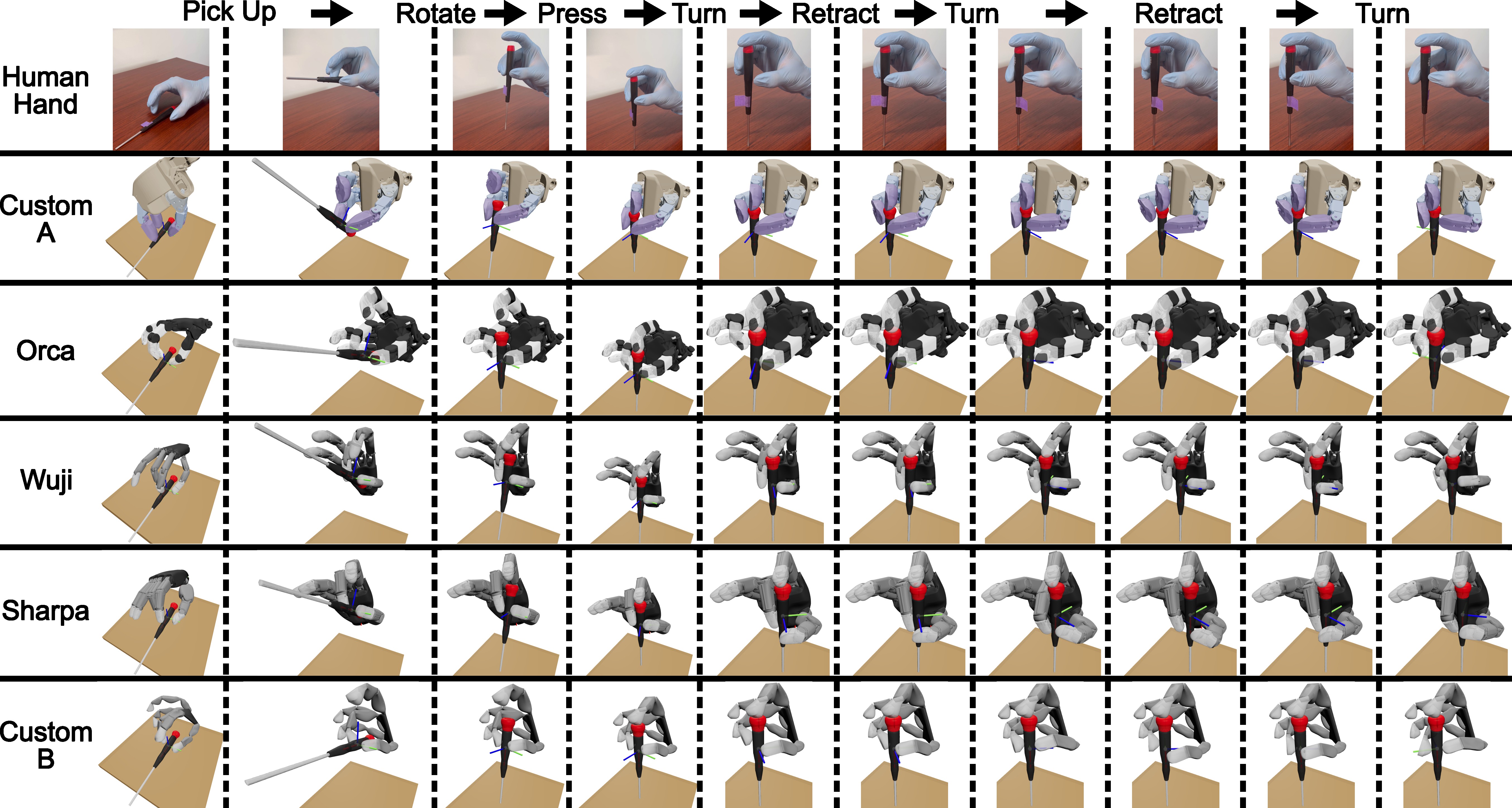}
  \captionof{figure}{Five different robotic hands picking up, loading, and wielding a precision screwdriver \cite{craftsman_precision_screwdriver_cad} in Drake simulator \cite{drake}, along with the human hand performing the same task as a visual reference.}
  \label{fig:headline}
  }
\makeatother
\maketitle
\addtocounter{figure}{-1}

\maketitle
\thispagestyle{empty}
\pagestyle{empty}

\begin{abstract}

This work investigates how to enable general multi-finger robotic hands to perform a complete tool manipulation process, which entails picking up a tool, loading it into a suitable pose, and then wielding it. Our approach is to consider the hand and the tool together as a hand-object mechanism (HOM) composed of sub-assemblies (SAs). Specifically, we formulate HOM and SA as a general and efficient representation of dexterous tool manipulation, allowing hand-object relative motions to be described by the same set of physically meaningful Cartesian relative contact velocities, irrespective of the hand's and the tool's kinematics and geometries. To demonstrate the HOM-SA representation's efficacy, we implemented it as a lightweight motion planning and contact estimation framework. Then, we tested the framework in simulation on five morphologically different robotic hands. The results show that the framework enabled all five hands to execute the complete tool manipulation process, achieving fine, dexterous behaviors even when driven by identical, simple reference trajectories. Furthermore, the results showcase the HOM-SA representation's generalizability across different hands, tools, and tasks, which is rooted in its kinematic and geometric foundation. Please visit the following website for videos: \href{https://sunyuw.github.io/hom-sim/}{https://sunyuw.github.io/hom-sim/}.

\end{abstract}
\section{Introduction}




Tool manipulation is an essential skill for humanoid robots to master. A defining characteristic of tool manipulation is tool-wielding, which centers on the particular ways to grasp and move a tool to exploit its function. Importantly, tool-wielding grasps and movements generally differ from how the same tool would be grasped and moved in pick-and-place \cite{sunyu-foundational_pose, contactgrasp}. Hence, for humanoid robots equipped with multi-finger hands, a complete tool manipulation process includes three phases: 1) picking up, 2) loading, and 3) wielding the tool. Loading means configuring the tool from the pose in which it is picked up to the pose in which it is ready to be wielded. In this work, we investigate how to enable general multi-finger robotic hands to achieve the entire pick-load-wield tool manipulation process. 

Our investigation begins with a mechanical intuition derived from human tool manipulation: the human hand configures itself and the tool into various parallel mechanisms to achieve provisional objectives in each phase of the tool manipulation process. These hand-object mechanisms (HOMs) are often simple, effective, and contain common sub-assemblies (SAs), much like the SAs of machines. 

Following this intuition, we identified specific SAs in human tool manipulation, and formulated the concepts of HOM and SA for general multi-finger robotic hands. This formulation establishes HOM and SA as a general and efficient representation of dexterous tool manipulation. It allows hand-object relative motions to be described by the same set of physically meaningful Cartesian relative contact velocities, irrespective of the hand's and the tool's kinematics and geometries. 

To demonstrate the HOM-SA representation's efficacy, we implemented it as a lightweight motion planning and contact estimation framework based on least squares and a complementary filter. Then, we tested the framework in simulation on five morphologically different robotic hands. The results show that the framework enabled all five hands to achieve the complete tool manipulation process with fine, dexterous behaviors, even when driven by identical, simple reference trajectories. Moreover, because of its kinematic and geometric foundation, the HOM-SA representation has little reliance on extensive data collection or model training. Setting up a new hand, tool, or task is straightforward and can be completed in a matter of hours. Lastly, we discuss the HOM-SA representation's limitations and considerations for implementing it on real-world hardware. 


\section{Related Works}



\subsection{Hand-Object Mechanism (HOM)}

With the established theories of rigid body kinematics, treating a multi-finger hand and the manipulated object as one parallel mechanism has been a common approach. Prior works adopted this approach to analyze the workspace properties of multi-finger precision grasps \cite{odhner_stable_precision_manipulation_with_underactuated_hands, rojas_gross_motion_analysis, julia_analyze_hands_with_parallel_robots} and to design robotic hands \cite{stewart_hand_1, sphinx_hand_1, hand_as_optimal_spherical_mechanism_2021, gripper_from_spherical_mechanism_2009}. 

However, in the workspace analysis works, HOM mostly serves as a qualitative inspiration without concrete definitions. In the hand design works, the HOMs correspond to the chosen parallel mechanism designs, so the results tend to be specific to these morphologies. The question of what a general HOM is and how to use it to control general multi-finger robotic hands remains unanswered. This work provides an answer by formulating the HOM-SA representation, concretely defining a HOM and the SAs that constitute it. 

\subsection{Human Hand Manipulation Studies}

Given the human hand morphology, a HOM and its SAs fundamentally reflect human manipulation behaviors, which have received extensive study. The study most relevant to this work is \cite{elliott_connolly_taxonomy}, as it contains types of precision in-hand manipulation that conceptually align with what we identified as SAs. Meanwhile, the concept of SA resonates with postural synergies \cite{human_grasp_synergies_1, human_grasp_synergies_2, human_grasp_synergies_3}, and the SAs we identified often operate on the basis of opposition and virtual finger \cite{iberall_opposition_space, arbib_virtual_finger}, which matches the behavior types in \cite{iberall_prehension_taxonomy}. 

Despite the conceptual overlap, this work differs from these human hand manipulation studies in approach and objective. These studies adopted an analysis approach, aiming to identify, classify, and elaborate on human hand manipulations as systematically and comprehensively as possible. By contrast, this work adopts a synthesis approach. Our objective is to enable general multi-finger robotic hands---which are often designed to mimic the human hand, but still radically differ from the human hand---to achieve the entire pick-load-wield tool manipulation process. Consequently, rather than meticulously analyzing human hand behaviors, this work focuses on formulating the HOM-SA representation and demonstrating the capabilities it unlocks.

\subsection{Robotic Manipulation Motion Planning and Control}

Three technical approaches are popular among recent works on robotic manipulation motion planning and control: 1) imitation learning, 2) reinforcement learning, and 3) optimal control. All three approaches tackle manipulation as an optimization problem based on certain models and policies. Their main difference is the types of models and policies used \cite{xianyi_humanoid_survey, kroemer_learning_for_manipulation_review}. This work aligns most closely with the optimal control approach \cite{sunyu-geodesic, sunyu-foundational_pose, rolling_contact_motion_planning-1, rolling_contact_motion_planning-2, xianyi-contact_mode_guided_motion_planning_3d}, and is an extension of \cite{sunyu-foundational_pose, sunyu-geodesic} from kinematic tool-wielding to the entire pick-load-wield tool manipulation process in the dynamic case. 

Moreover, many recent manipulation motion planning and control works have adopted a top-down perspective. They aim to develop one end-to-end model and/or policy with sufficient generalization capacity to encompass all tasks that the robot is expected to encounter. While these works have achieved remarkable results, their heavy reliance on data and computational resources can restrict their accessibility and ease of deployment. By contrast, this work adopts a bottom-up perspective based on the concepts of HOM and SA. Hence, the resultant framework is lightweight and easier to set up for different robots and tasks.

\section{Methods}

\subsection{Mechanical Intuition and Identification of Sub-Assemblies from Human Tool Manipulation}

The essence of tool manipulation is to achieve some desired Cartesian motion of, and/or wrench applied by, the tool. If the tool is mounted on a mechanical mechanism that admits the desired motion and wrench, executing the tool manipulation will be straightforward. 

From a mechanical design perspective, it is challenging to design a single mechanism that accounts for the immense variety and uncertainties of tools and tasks. However, we can apply this perspective to interpret human tool manipulation, while borrowing ideas from artificial intelligence: the human hand acts as a ``mechanism approximator", much like an artificial neural network (ANN) acts as a function approximator. With many neurons, an ANN can approximate many functions \cite{universal_function_approximation_theorem}. Analogously, with many degrees of freedom (DoFs), the human hand can approximate many mechanical mechanisms, and is effectively forming the best mechanism it can with the tool to achieve the desired motion and wrench in each phase of the pick-load-wield tool manipulation process. 

Furthermore, many contemporary ANNs operate based on tokens, instead of individual numbers or characters. Analogously, evidence suggests that the human hand operates based on postural synergies and grasp and movement types, instead of individual DoFs \cite{human_grasp_synergies_1, human_grasp_synergies_2, human_grasp_synergies_3, elliott_connolly_taxonomy, iberall_prehension_taxonomy}. Hence, we seek to identify the ``tokens" of human tool manipulation, i.e., the atomic units that constitute the HOMs used by humans for tool manipulation. We term these atomic units ``sub-assemblies (SAs)"---akin to the SAs of a machine---to reflect their mechanical origin. 

A common SA is a two-finger antipodal grasp, or pinch. A pinch as a SA is more than a static posture; it can change its function by regulating the contact positions and forces. When the two contacts are directly opposed and the contact forces are large, the pinch acts as a fixed joint, immobilizing the grasped tool. When the forces are reduced but sufficient to hold the tool, the pinch acts as a revolute or prismatic joint, allowing the tool to rotate or translate in-hand. When the two contacts are offset from each other, the pinch acts as an actuator to axially rotate the tool, similar to the racks of the double rack-and-pinion mechanism. 

Another common SA is a single contact between one finger and the tool, which we call a ``pusher". By controlling the finger's motion and contact force, a pusher can act as an actuator or a supporting structure. Frequently, a pusher pairs with a pinch and acts as an actuator that rotates or translates the tool about or along the joint formed by the pinch. Sometimes, a pusher pushes the tool against an environmental contact, forming a pinch with just one finger. 



\subsection{Definitions of Hand-Object Mechanism and Sub-Assembly for General Multi-Finger Robotic Hands}

Next, we formulate the HOM-SA representation for general multi-finger robotic hands. Assuming nominally rigid bodies and that the kinematics and geometries of a hand and an object are known, we define a HOM to consist of the hand, the object, and the generalized contact frames (GCFs). 

\begin{figure}[ht]
    \centering
    \includegraphics[width=0.485 \textwidth]{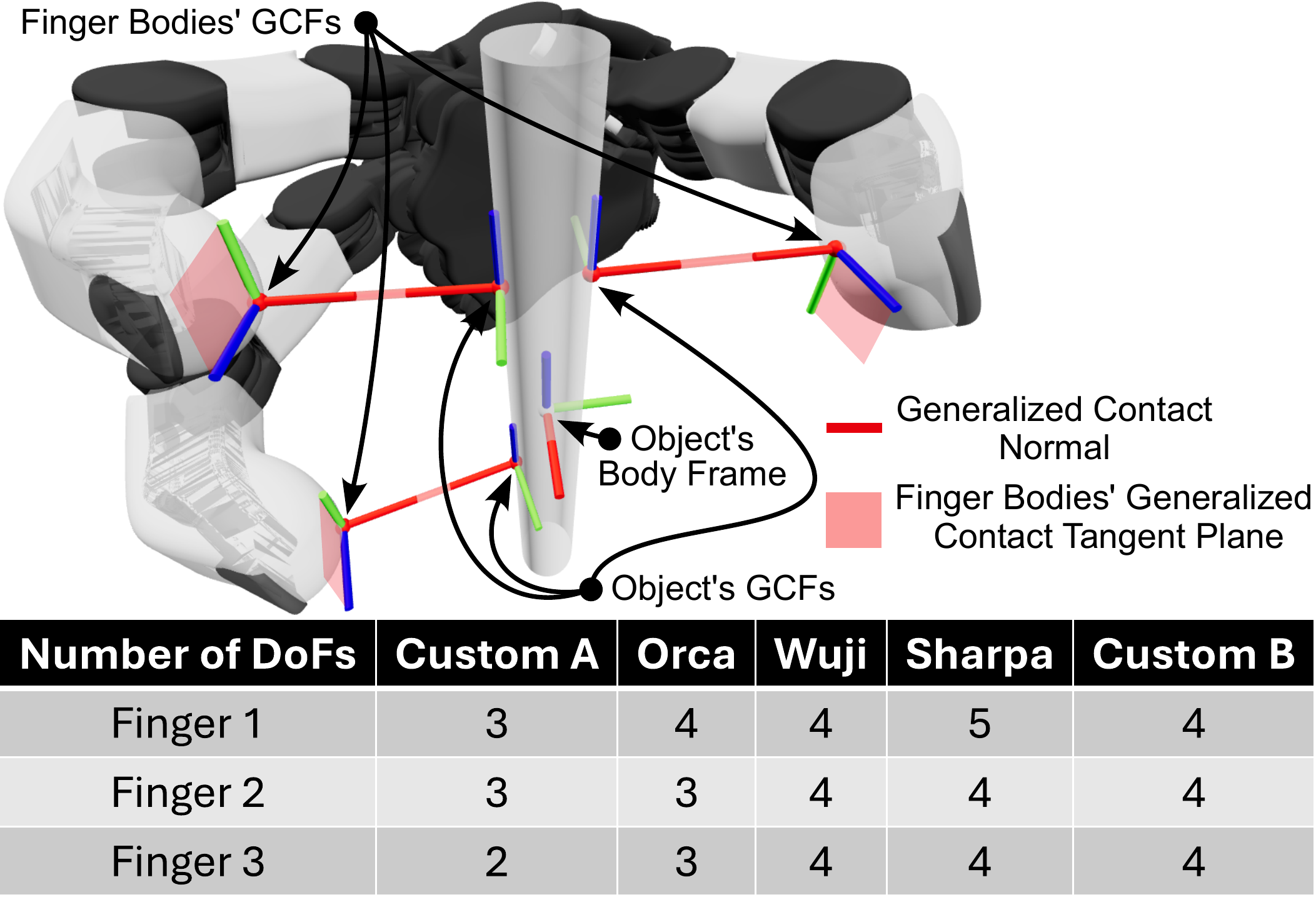}
    \caption{(Top) The hand-object mechanism consisting of a three-finger Orca hand \cite{orcahand_description}, a primitive cylinder, and the generalized contact frames (GCFs), when the fingers and the object are not in contact. The GCFs' $x$-axes are the normal axes, and their $y$-axes are maximally aligned with each finger body's joint axis. The axes' color convention is red-green-blue $\rightarrow x, y, z$. (Bottom) Each finger's number of DoFs for the five robotic hands used in the experiments. Fingers $4$--$5$ are ignored as they were kept still in the experiments. }
    \label{fig:gcf_schematic}
\end{figure}

Specifically, we define GCFs as two time-updated Cartesian frames for each finger, one fixed to the body of the finger expected to contact the object---typically the distal link---and one fixed to the object. Then, we select one of the three Cartesian axes as the GCFs' normal axes. When the finger body and the object are not in contact, the GCFs' origins are the respective bodies' witness points, i.e., the points on each body's surface closest to the other body. The GCFs' normal axes align with the gradients of the signed distance function between the two bodies. When the finger body and the object are in contact, the GCFs' origins are the points on the respective bodies' surfaces closest to the true instantaneous contact point. The GCFs' normal axes are the unit vectors pointing from the contact point to the respective GCFs' origins, which always point outward relative to the respective bodies' surfaces unless the finger body and the object contact at precisely one point without interpenetration. In this edge case, the GCFs' normal axes are the outward-pointing surface normals at the respective GCFs' origins. Fig. \ref{fig:gcf_schematic} (top) shows an example of a HOM and its GCFs.

The GCFs kinematically connect the hand and the object into a HOM, regardless of whether the fingers and the object are in contact. Mathematically, let $\{ P \}, \{ B \}, \{ L \}$ denote the palm's body frame, a finger body's or the object's body frame, and a GCF, respectively. Let $T_{P B} \in SE \left( 3 \right)$ denote $\{ B \}$'s pose relative to $\{ P \}$ as a $4 \times 4$ homogeneous transformation matrix, let $V^{b}_{P B} \in \mathbb{R}^{6}$ denote $\{ B \}$'s body twist relative to $\{ P \}$, and let $\text{Ad}$ denote the adjoint operator. Let subscripts $0$ and $1$ denote frames attached to the object and the finger body, respectively. For each finger in a HOM, since its GCF is fixed to the finger body, 
\begin{align}
    V^{b}_{L_{0} L_{1}} = \text{Ad}_{T_{L_{1} B_{1}}} V^{b}_{P B_{1}} -\text{Ad}_{T_{L_{1} B_{0}}} V^{b}_{P B_{0}} \text{\cite{sunyu-geodesic}}.
    \label{eq:relative_contact_twist}
\end{align}

The (\ref{eq:relative_contact_twist})'s of all fingers constitute the non-holonomic constraints that fully specify the HOM. The twist $V^{b}_{L_{0} L_{1}}$ contains each finger's four relative contact velocities: 1) $v_{roll} \in \mathbb{R}^{2}$, the angular velocity in the generalized contact tangent plane, i.e., the plane defined by the GCF's two non-normal axes, 2) $v_{spin} \in \mathbb{R}$, the angular velocity about the GCF's normal axis, 3) $v_{slide} \in \mathbb{R}^{2}$, the linear velocity in the generalized contact tangent plane, and 4) $v_{sep} \in \mathbb{R}$, the linear velocity along the GCF's normal axis. $v_{roll}, v_{spin}, v_{slide}$, and $v_{sep}$ respectively describe the finger body's rolling, spinning, sliding, and separation/approach motions relative to the object's surface, and are simply the elements of $V^{b}_{L_{0} L_{1}}$. For example, if the GCF's normal axis is the $x$-axis, $V^{b}_{L_{0} L_{1}} = \begin{bmatrix} v_{spin}^{\intercal} & v_{roll}^{\intercal} & v_{sep}^{\intercal} & v_{slide}^{\intercal} \end{bmatrix}^{\intercal}$. 

Lastly, we define a SA as a set of constraints that couple particular fingers' relative contact velocities and contact forces. Specifically, a pinch SA constrains two fingers' $v_{sep}$ terms to be identical, so the two fingers move toward or away from the object simultaneously, if they are not in contact with the object. When in contact, a pinch constrains two fingers' contact forces to have the same magnitude and opposite directions, applying a zero net wrench to the object and achieving force closure if the line connecting the contact points is within their friction cones \cite{mls_1994}.


\subsection{Motion Planning with Relative Contact Velocities}

To plan each finger's motion, we substitute the finger body's Jacobian equation $V^{b}_{P B_{1}} = J^{b}_{P B_{1}} u$ into (\ref{eq:relative_contact_twist}), where $u$ denotes the finger's joint velocities. Then, we treat the object's twist $V^{b}_{P B_{0}}$ as a reference input, and solve (\ref{eq:relative_contact_twist}) for the finger's reference joint velocities, $u_{ref}$, via least squares of the relative contact velocities, while respecting joint position and velocity limits
\begin{align}
u_{ref} = \argmin_{u} & \Big( \sum_{i} w_{i} \lVert v_{i, ref} - v_{i} ( u ) \rVert^2 + w_{damp} \lVert u \rVert^2 \notag \\
& + w_{healthy} \lVert k_p \left( \theta_{healthy} - \theta \right) - u \rVert^2 \Big) \label{eq:motion_planner} \\
\text{s.t.} \quad \theta_{min} \le \theta + &\Delta t u \le \theta_{max} \text{ and } u_{min} \le u \le u_{max}, \notag
\end{align}
where $i \in \{ roll, spin, slide, sep \}$. $v_{i, ref}$ is the reference for each relative contact velocity, $\theta$ is the current joint positions, $\Delta t$ is the update time period, and the $w$'s are weights with $w_{healthy} \ll w_{damp} \ll w_{i}$. The $w_{damp} \lVert u \rVert^2$ term is a damped least-squares regularizer to stabilize the reference joint velocities near singularities. The last cost term is a regularizer to gently pull the finger close to a pre-recorded healthy pose, $\theta_{healthy}$. This prevents a finger with redundant DoFs from being trapped in regions of its joint space that are difficult to exit. After obtaining $u_{ref}$, we time-integrate $u_{ref}$ into reference joint positions $\theta_{ref}$, and compute the finger's joint torques to track $\theta_{ref}$ and $u_{ref}$ using an inverse dynamics controller. To address integral windup, we clamp $v_{sep, ref}$ to zero when a finger body is within a small distance to the object, preventing the motion planner from commanding the finger to penetrate the object. 

Each finger's motion can be commanded via either the four reference relative contact velocities, $v_{i, ref}$, or the reference object body twist, $V^{b}_{P B_{0}}$, which we set differently based on the finger's SA. Through $V^{b}_{P B_{0}}$, the fingers can manipulate the object as the HOM's effector and achieve rich dexterous behaviors with simple commands, irrespective of the fingers' kinematics and geometries. 

Specifically, the two fingers in a pinch SA will have the same $V^{b}_{P B_{0}}$. When a third finger contacts the object and acts as a pusher to rotate the object about the pinch axis, the pinching fingers will have $V^{b}_{P B_{0}} = 0$, whereas the pusher finger will have
\begin{align}
    V^{b}_{P B_{0}} = \text{Ad}_{T_{B_{0} M}} \underbrace{ \begin{bmatrix} \begin{bmatrix} s & 0 & 0 \end{bmatrix}^{\intercal} & 0_{1 \times 3} \end{bmatrix}^{\intercal} }_{V^b_{P M}},
    \label{eq:pin_pusher_refobject_twist}
\end{align}
where $s \in \mathbb{R}$ is the reduced-order reference speed, and $\{ M \}$ is the pinch frame fixed to the object, whose origin is the midpoint of the line segment connecting the two pinching fingers' contact points and whose $x$-axis is parallel to the line segment. This enables control of the object's rotation about the pinch axis via a single scalar, $s$, as if the pinch is a revolute joint and the pusher is the revolute joint's actuator. 


\subsection{Contact Estimation for Stable Contact Force Control}

Besides motion, the HOM-SA representation involves contact force control for each finger, which requires an action point and a direction. In general, the instantaneous contact point (ICP) is a noisy measurement sensitive to short-term contact movement, and is unsuited to be the action point in its raw form. To achieve stable contact force control, we pass the ICP into a complementary filter to estimate the action point as a continuous state using a process model based on roll-slide contact kinematics and geodesic tracing \cite{sunyu-geodesic}. 

Specifically, let $\{ C_0 \}$ and $\{ C_1 \}$ be Cartesian frames {\it moving} on the object's and the finger body's respective surfaces. Define their normal axes to be the respective bodies' outward-pointing surface normals. Then, 
\begin{align}
    &p_{B_{j} C_{j}} \left[ t + 1 \right] = \textit{trace\_geodesic} \left( p_{B_{j} C_{j}} \left[ t \right], \dot{g}_{j} \Delta t \right) \label{eq:geodesic_tracer} \\
    &\dot{g}_{j} = \dot{g}_{j, ref} + k_j E \left( p_{B_j L_j} \left[ t \right] - p_{B_j C_j} \left[ t \right] \right) \\
    &k_j = k_{min} + ( k_{max}- k_{min} ) \left( 1 - e^{ -3 \alpha^{-1} \lVert \dot{g}_{j, ref} \rVert } \right),
\end{align}
where $j \in \{ 0, 1 \}$, $p_{B_j C_j} \in \mathbb{R}^3$ is $\{ C_j \}$'s Cartesian position relative to $\{ B_j \}$, $\dot{g}_{j} \in \mathbb{R}^{2}$ is the geodesic velocity in $\{ C_{j} \}$'s tangent plane at discrete time step $t$, $\dot{g}_{j, ref}$ is the reference for $\dot{g}_{j}$, and $E \in \mathbb{R}^{2 \times 3}$ is a matrix of $0$s and $1$s extracting the non-normal elements of a Cartesian position. $k_j \in \mathbb{R}^{+}$ is the filter gain. $\alpha \in \mathbb{R}$ is the geodesic speed threshold above which $k_j$ exceeds $95\%$ of its maximum value, as $\left( 1 - e^{-3} \right) \approx 0.95$. Since $\{ L_j \}$---which represents the ICP---and $\{ C_j \}$ are both on the finger body's or the object's surface, their origins will converge when $\lVert \dot{g}_{j, ref} \rVert$ is small.  

To obtain the reference geodesic velocities $\dot{g}_{j, ref}$, we solve the roll-slide contact kinematics equations without the normal elements \cite{sunyu-geodesic}
\begin{align}
    S \begin{bmatrix} \text{Ad}_{T_{C_1 C_0}} G_0 \left( K_0 \right) & -G_1 \left( K_1 \right) \end{bmatrix} 
    \begin{bmatrix} \dot{g}_{0, ref} \\ \dot{g}_{1, ref} \end{bmatrix} = S V^{b}_{L_0 L_1, ref}. \notag
\end{align}
Here, $G_j \left( K_j \right) \in \mathbb{R}^{6 \times 2}$ is the object's or the finger body's surface Jacobian constructed from the curvature tensor $K_j \in \mathbb{R}^{2 \times 2}$ at $p_{B_j C_j} \left[ t \right]$, $S \in \mathbb{R}^{4 \times 6 }$ is a matrix of $0$s and $1$s extracting the non-normal elements of a twist, and $S V^{b}_{L_0 L_1, ref} = \begin{bmatrix} v_{roll, ref}^{\intercal} & v_{slide, ref}^{\intercal} \end{bmatrix}^{\intercal}$. 

This contact estimator allows fine time integration of $\{ C_j \}$ when $\lVert \dot{g}_{j, ref} \rVert$ is small to smooth $\{ C_j \}$'s movement, while ensuring that $\{ C_j \}$ remains close to the ICP. With $\{ C_1 \}$'s origin and normal axis as the action point and direction, respectively, stable contact force control can be achieved. 

\subsection{Software Implementation and Experimental Setup}

We implemented the motion planner and the contact estimator as a unified framework to actualize the HOM-SA representation in Drake \cite{drake}. We chose Drake for its hydroelastic contact model, as the pinch SA's operation requires frictional torque, which point contact models struggle to handle. To enable hydroelastic contact, we tetrahedralized the triangular surface meshes of the bodies expected to make contact using Tetgen \cite{tetgen_original_paper}. Then, we selected the $x$-axis as the GCFs' normal axes, computed the hydroelastic contact surfaces' centers of pressure as the ICPs, and obtained the GCFs' origins and normals from signed distance and closest point queries between the contact points and the convex decompositions of the finger bodies' and the object's surface meshes \cite{fcpw, coacd}. The tangent axes of the finger bodies' GCFs have their $y$-axes maximally aligned with the respective finger body's joint axis, so absolute directions are maintained in the generalized contact tangent planes. 

Since a convex-decomposed body may contain multiple meshes and a geodesic may traverse them, we implemented an approximate geodesic tracer for the contact estimator by tracing an arc of the osculating circle. This approximation is close to true geodesic tracing when the geodesic traced is short, which is generally our case. The osculating circle is computed based on vertex-interpolated principal curvature values and directions \cite{igl}. 

To provide reference commands to the HOM-SA framework for experimental validation, we built a graphical user interface (GUI) containing eight control sliders for each finger: 1) SA selector, 2) contact force magnitude, and 3)--8) the six scalars in the four reference relative contact velocities in (\ref{eq:motion_planner}). Additionally, the GUI has six sliders for the reference object body twist, i.e., the $V^{b}_{P B_{0}}$ in (\ref{eq:relative_contact_twist}), one slider for the object's reduced-order reference speed, i.e., the $s$ in (\ref{eq:pin_pusher_refobject_twist}), and six sliders for the palm's twist, as we control the palm as a fully actuated floating base. We ran the simulation at $1$ kHz and the framework at $500$ Hz. The visualization updated at under $20\%$ of real-time speed, giving the human operator sufficient time to determine the appropriate high-level reference commands and react to the hand's behavior. 

We tested the HOM-SA framework on five morphologically different robotic hands in four experiments. The hands are: 1) Orca hand V2 \cite{orcahand_description}, 2) Wuji hand \cite{wuji_description}, 3) Sharpa Wave \cite{sharpa_wave_urdf}, 4) custom A, a three-finger hand designed for power grasps, and 5) custom B, a five-finger hand designed based on the human hand's anatomy. Fig. \ref{fig:gcf_schematic} (bottom) shows these hands' numbers of DoFs per finger, with the thumb through pinky denoted as fingers 1--5, respectively. The weights in (\ref{eq:motion_planner}) are the same for all hands. The next section will detail the four experiments. 
\section{Results and Discussion}

\subsection{Moving Around a Cylinder Without Contact}

\begin{figure*}[ht]
    \centering
    \includegraphics[width=0.99 \textwidth]{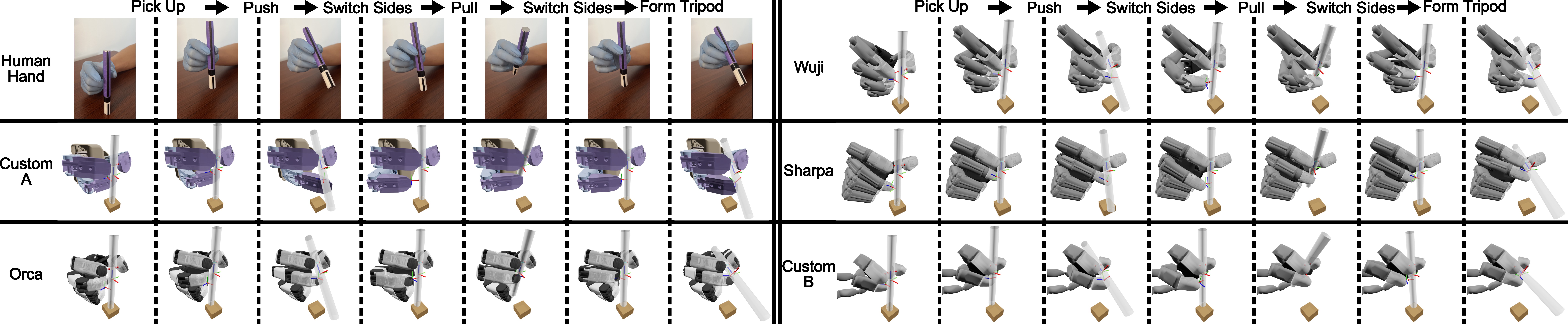}
    \caption{Snapshots of the experiment of picking up a cylinder, rotating it about the pinch axis via control of the single scalar $s$ in (\ref{eq:pin_pusher_refobject_twist}), and forming a tripod grasp, along with the human hand performing the same task as a visual reference.}
    \label{fig:pin_pusher_snapshots}
\end{figure*}

In our first experiment, a cylinder stood on a table, and each hand was initialized in a pregrasp pose close to pinching the cylinder with fingers 1 and 2, as shown in the first column for each hand in Fig. \ref{fig:pin_pusher_snapshots}. Then, via the GUI, we let finger 2 move around the cylinder without contact by sequentially commanding a step trajectory to each scalar element of the reference relative contact velocities, while keeping all other elements at zero. The step magnitudes were $40$ deg/s and $20$ mm/s for angular and linear velocities, respectively. 

\begin{figure*}[ht]
    \centering
    \includegraphics[width=0.99 \textwidth]{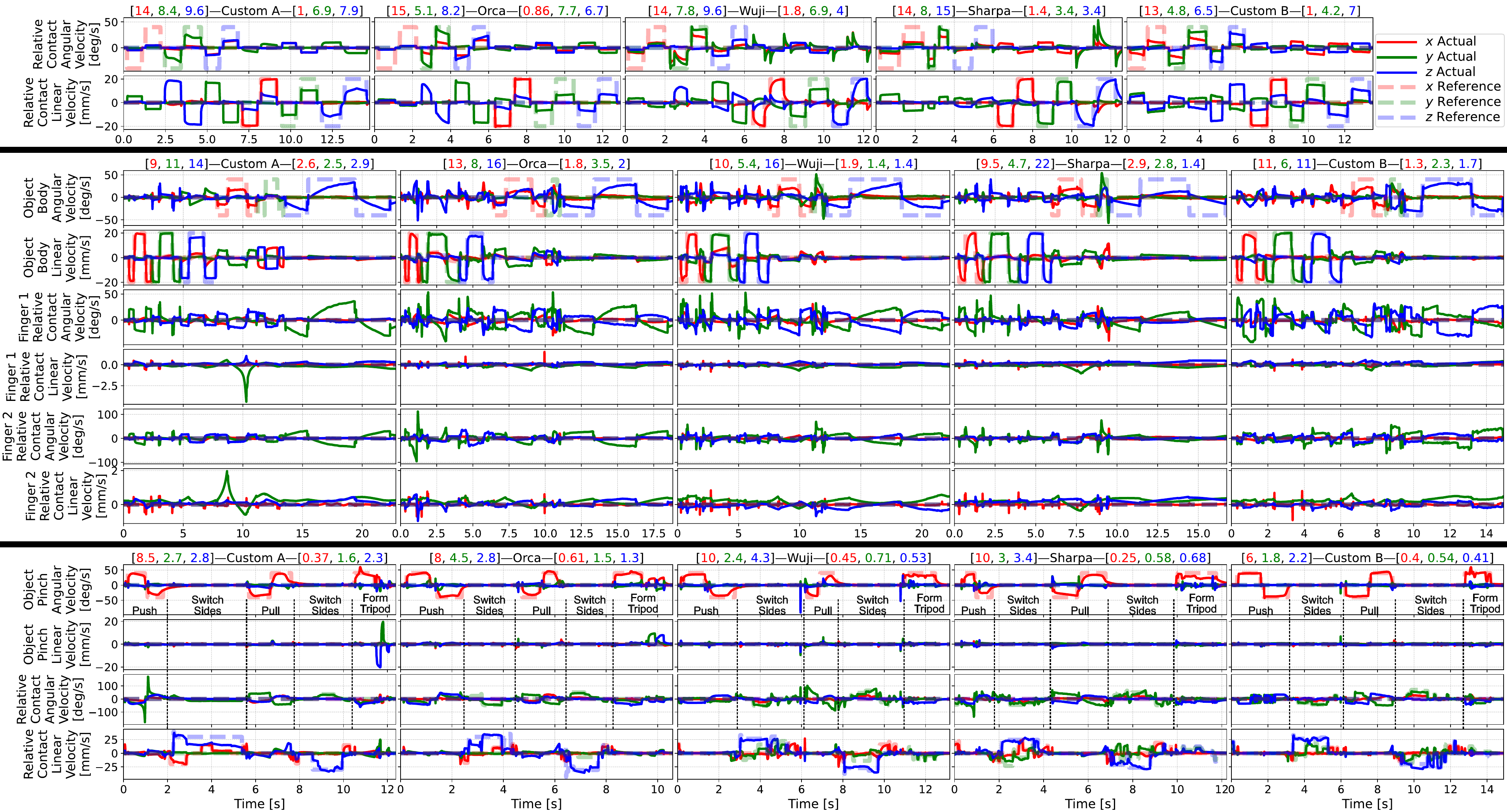}
    \caption{Trajectories from the three quantitative experiments. (Top block) Moving finger 2 around a cylinder without contact. (Middle block) Picking up and manipulating a cylinder based on its body twist. (Bottom block) Rotating a cylinder about the pinch axis and forming a tripod grasp. The 3-vectors on the left and right of each hand's name next to the ``---" are the root mean squared errors (RMSEs) for angular and linear velocity tracking, respectively, over the entire trajectory. The RMSEs in the top, middle, and bottom blocks are for relative contact velocities, object body velocities, and object pinch velocities, respectively. In each block, the plots on the same row and column share the same vertical and horizontal ticks and scales, respectively. All plots' horizontal axes represent time. }
    \label{fig:quantitative_experiments}
\end{figure*}

Fig. \ref{fig:quantitative_experiments} (top) shows the results. Despite the hands' different morphologies, the HOM-SA framework enabled all hands to achieve similar Cartesian trajectories with consistent tracking performance in terms of the ratio of the root mean squared errors to the commanded step magnitudes. The $y$-axis angular velocity and $x$-axis linear velocity were tracked most closely. This is because the GCF's $y$-axis is maximally aligned with the finger body's joint axis, and the GCF's $x$-axis is the normal axis. Hence, these reference velocities cleanly correspond to the finger's flexion and extension. Conversely, most hands exhibited the worst tracking performance for the $x$-axis angular velocity. This is because the finger has fewer than six DoFs and cannot satisfy all reference velocities simultaneously. To satisfy the non-zero reference $x$-axis angular velocity, the finger must adduct or abduct. However, any adduction or abduction would induce non-zero velocities in other axes, which are penalized. 

Sometimes, this type of coupling between angular and linear velocities can facilitate the finger's movement rather than hindering it. For example, translation along the GCF's $z$-axis induces rotation about the $y$-axis. Hence, a single non-zero $z$-axis reference linear velocity enables the finger to trace the cylinder's surface at a constant distance and switch sides relative to the cylinder, as the finger moves into and out of the space between the cylinder and the palm. 

\subsection{Picking Up and Manipulating a Cylinder as Effector}

In our second experiment, each hand picked up the cylinder in a pinch and manipulated it in-hand as the HOM's effector based on simple step references. Specifically, we selected the pinch SA for fingers 1 and 2, and moved them toward the cylinder by commanding a positive $v_{sep, ref}$. With the anti-integral windup, the fingers stopped just before contacting the cylinder. Then, we commanded a constant pinching force and moved the palm up, letting the fingers grasp the cylinder and lift it off the table, as the second column for each hand in Fig. \ref{fig:pin_pusher_snapshots} shows. Next, we set the weights $w_{roll}$ in (\ref{eq:motion_planner}) to zero to allow rolling, and commanded a step trajectory to each scalar element of the reference object body twist, the $V^{b}_{PB_{0}}$ in (\ref{eq:relative_contact_twist}), with the same angular and linear magnitudes as in the first experiment, while keeping all other elements at zero. The reference relative contact velocities were also set to zero. The fingers translated and rotated the cylinder along and about each axis of the cylinder's body frame by rolling on the cylinder's surface, while minimizing sliding and spinning to facilitate a stable grasp. 

Fig. \ref{fig:quantitative_experiments} (middle) shows the results. The HOM-SA framework enabled all hands to manipulate the cylinder as the HOM's effector, with reasonable tracking performance for object body linear velocity. For angular velocity, the $x$- and $z$-trajectories mostly follow the step references but have smaller magnitudes. Tracking of the $y$-axis angular velocity appears worse and shows noticeable crosstalk in translations along the $x$- and $z$-axes. This is because the cylinder's body frame $y$-axis is nearly parallel to the pinch axis, and the contact points themselves cannot rotate the cylinder. To satisfy a non-zero $y$-axis reference angular velocity while keeping other velocity elements at zero, the fingers compensated by translating the cylinder along the $x$- and $z$-axes. 

When moving the cylinder, all hands exhibited noticeable relative contact angular velocities about the $y$- and $z$-axes, but small relative contact linear velocities along these axes. This means that the fingers maintained a stable grasp by rolling on the cylinder's surface with little sliding. Moreover, since the cylinder's body frame $z$-axis is its longitudinal axis, the simple step reference of the $z$-axis angular velocity yielded the fingers' dexterous twisting motions for all hands, similar to how a human hand turns a screwdriver. This demonstrates the HOM-SA framework's effectiveness in generating dexterous behaviors from even identical, simple commands for different hands. 

\subsection{Rotating a Cylinder About the Pinch Axis and Forming a Tripod Grasp}

In our third experiment, each hand picked up the cylinder with fingers 1 and 2 in a pinch to form a revolute joint. Then, finger 3 entered a pusher SA and rotated the cylinder about the pinch axis in both directions. This motion is commonly used by humans to load a tool.

Specifically, we let finger 3 push and rotate the cylinder away from the palm by commanding a step reference trajectory of $40$ deg/s to the reduced-order reference speed $s$ in (\ref{eq:pin_pusher_refobject_twist}). After finger 3 was almost fully extended, we negated the step reference; finger 3 maintained contact with the cylinder as the cylinder rotated back to a vertical pose by gravity. Then, via the reference relative contact velocities, we let finger 3 break contact, switch sides relative to the cylinder, make contact, and repeat the rotation cycle by pulling the cylinder toward the palm. Lastly, finger 3 switched sides, pushed the cylinder away from the palm again, and coordinated with fingers 1 and 2 to form a tripod grasp of the cylinder. Fig. \ref{fig:pin_pusher_snapshots} shows the entire process. All other setups were identical to the second experiment, except that the weights $w_{spin}$ in (\ref{eq:motion_planner}) were set to zero for the two pinching fingers to allow spinning. 

Fig. \ref{fig:quantitative_experiments} (bottom) shows the quantitative results. The object pinch velocities refer to the velocities in the $V^b_{P M}$ term in (\ref{eq:pin_pusher_refobject_twist}), which should have all elements at zero except the $x$-axis angular element if the cylinder purely rotates about the pinch axis. The plotted relative contact velocities are for finger 3, the pusher finger. All hands achieved reasonable tracking of the reference object pinch velocities, while exhibiting small relative contact linear velocities when the cylinder was rotating. This means that finger 3 acted as the actuator of the revolute joint formed by the pinch, and mainly rolled on the cylinder with little sliding during the rotation cycles. This result was accomplished via the simple step references of the single scalar $s$ in (\ref{eq:pin_pusher_refobject_twist}), again demonstrating the HOM-SA framework's effectiveness in generating dexterous behaviors from even identical, simple commands for different hands. 

\subsection{Picking Up, Loading, and Wielding a Screwdriver}

In our fourth experiment, we showcase the HOM-SA framework by letting each hand pick up, load, and wield a precision screwdriver with a non-convex geometry \cite{craftsman_precision_screwdriver_cad}. Fig. \ref{fig:headline} shows this demonstration. Specifically, fingers 1 and 3 picked up the screwdriver from a table in a pinch with a pinching force large enough to immobilize it in-hand. After lifting the screwdriver clear of the table, fingers 1 and 3 reduced their pinching force, allowing the screwdriver to rotate about the pinch axis to a vertical pose by gravity. Then, finger 2 contacted the screwdriver and rotated it about the pinch axis to align it perpendicular to the table via the reduced-order reference speed $s$ in (\ref{eq:pin_pusher_refobject_twist}); the palm moved down until the screwdriver's tip contacted the table. Next, finger 2 pressed the screwdriver against the table, stabilizing it; fingers 1 and 3 turned the screwdriver counterclockwise via the reference object body angular velocity about the screwdriver's longitudinal axis. This first turn had a limited motion range since the two contacts were directly opposed. Thus, fingers 1 and 3 retracted to an offset pose ready for another turn by breaking contact and moving clockwise around the screwdriver. After the retraction, fingers 1 and 3 turned the screwdriver again. This retract-turn cycle repeated multiple times. 

This demonstration shows the level of dexterity that the HOM-SA framework can enable different hands to achieve. Moreover, since the framework requires only the hand's and the object's kinematics and geometries, setting up a new hand or object is straightforward. In fact, for custom B, the entire process from configuring the kinematic models and tetrahedralizing the meshes to completing the screwdriver demonstration took less than five hours. 

\subsection{Limitations and Future Work}

To generate high-quality manipulations with the HOM-SA framework, the hand must start from suitable pregrasp poses. This is because the constraints (\ref{eq:relative_contact_twist}) are non-holonomic; thus, the motion planner (\ref{eq:motion_planner}) is inherently velocity-based. In this work, we manually configured the hands into the pregrasp poses based on heuristics. However, reaching high-quality pregrasp poses is non-trivial in general. This limitation can be addressed by augmenting the HOM-SA framework with existing grasp synthesizers and poses from human hand grasp datasets \cite{grasp_synthesis_survey, contactgrasp, dexonomy, arctic_dataset}. 

For future work, we envision the HOM-SA representation as a new ``language" of dexterous manipulation that can be broadly leveraged in teleoperation and robot learning. With its generalizability and efficiency, this representation can serve as a unified state-action space for shared autonomy, reinforcement learning, and large behavior models, potentially enabling cross-embodiment teleoperation and more scalable model and policy training for dexterous manipulation. Furthermore, the relative contact velocity-based motion planner and contact estimator can be integrated into robot learning systems to guide reward shaping and fine-tuning. 

Lastly, we are currently implementing the HOM-SA framework on real-world hardware. The main challenge is contact sensing, which we are tackling via pose tracking \cite{foundationpose} and intrinsic tactile sensing \cite{intrinsic_tactile_sensing_original}. We plan to fuse the two sensing modalities into a pipeline similar to our current contact estimator to improve the overall sensing quality. 



\section{Conclusions}

This work presents the concepts of hand-object mechanism (HOM) and sub-assembly (SA) for general multi-finger robotic hands, and formulates them as a general and efficient representation of dexterous tool manipulation. At its technical core, the HOM-SA representation isolates the Cartesian normal and tangential relative velocities between each finger and the tool with the generalized contact frames. This allows the hand's motion to be commanded via a lightweight least-squares optimization of these physically meaningful relative contact velocities, which concretely describe the rolling, spinning, sliding, and separation/approach motions between the fingers and the object. Meanwhile, a complementary filter facilitates stable contact force control by fusing noisy contact point measurements into a process model based on roll-slide contact kinematics and geodesic tracing. 

We implemented the HOM-SA representation as a framework in simulation, and tested it on five morphologically different robotic hands. The results show that the framework enabled all five hands to manipulate a cylinder in a consistent manner as if each hand formed a parallel mechanism with it, achieving dexterous behaviors from even identical, simple reference trajectories. Then, we showcased the framework by enabling each hand to pick up, load, and wield a precision screwdriver. All five hands accomplished the task with fine, dexterous motions. These results demonstrate the HOM-SA representation's generalizability and effectiveness, motivating its broader applications and real-world implementation. 


\section*{Acknowledgment}

This research was partially supported by the RAI Institute, by NSF Convergence Accelerator award ITE-2344109, by the National Institute of Biomedical Imaging and Bioengineering of the National Institutes of Health under Award Number R01EB036842, and by the Institute of Information \& Communications Technology Planning \& Evaluation (IITP) grant funded by the Korean government (MSIT) (No.RS-2025-25463654).  


\bibliographystyle{IEEEtran}
\bibliography{references.bib}


\end{document}